\documentclass{IEEEtran}
\usepackage{cite}
\usepackage{amsmath,amssymb,amsfonts}
\usepackage{graphicx}
\usepackage{textcomp,nicefrac}
\usepackage{url}
\usepackage{hyperref}
\usepackage[sort&compress,square,numbers]{natbib}
\usepackage{todonotes}

\def\BibTeX{{\rm B\kern-.05em{\sc i\kern-.025em b}\kern-.08em
T\kern-.1667em\lower.7ex\hbox{E}\kern-.125emX}}
\begin{document}

\title{PGNAA Spectral Classification of Metal with Density Estimations}

\author{Helmand Shayan, Kai Krycki, Marco Doemeland, Markus Lange-Hegermann
\thanks{Manuscript received August 26, 2022; (Corresponding e-mail:
helmand.shayan@th-owl.de)}
\thanks{The MetalClass project is funded by the German Federal Ministry of Education and Research with the funding reference number 01IS20082B.}
\thanks{\textbf{Helmand Shayan} is with the inIT - Institute Industrial IT, OWL University of Applied Sciences and Arts in Lemgo, Germany (e-mail: helmand.shayan@th-owl.de).}
\thanks{\textbf{Kai Krycki} is with the Department of Mathematical Methods and Software Development, AiNT GmbH in Stolberg, Germany 
(e-mail: krycki@nuclear-training.de).}
\thanks{\textbf{Marco Doemeland} was with the Department of Mathematical Methods and Software Development, AiNT GmbH in Stolberg, Germany 
(e-mail: marcodoe@gmx.de).}
\thanks{\textbf{Markus Lange-Hegermann} is with the inIT - Institute Industrial IT, OWL University of Applied Sciences and Arts in Lemgo, Germany (e-mail: markus.lange-hegermann@th-owl.de).}
}

\maketitle

\begin{abstract}
For environmental, sustainable economic and political reasons, recycling processes are becoming increasingly
important, aiming at a much higher use of secondary raw materials. Currently, for the copper and aluminium
industries, no method for the non-destructive online analysis of heterogeneous materials are available. The
Prompt Gamma Neutron Activation Analysis (PGNAA) has the potential to overcome this challenge. A difficulty when using PGNAA for online classification arises from the small amount of noisy data, due to
short-term measurements. In this case, classical evaluation methods using detailed peak by peak analysis
fail. Therefore, we propose to view spectral data as probability distributions. Then, we can classify material
using maximum log-likelihood with respect to kernel density estimation and use discrete sampling to optimize hyperparameters. For measurements of pure aluminium alloys we achieve near perfect classification of
aluminium alloys under 0.25 second.
\end{abstract}


\section{Introduction}
\label{sec:introduction}
\IEEEPARstart{T}{he} recycling of scrap available in Europe as secondary raw
material is the safest, most sustainable and economical form
of raw material supply, which is available despite political
conflicts with mining countries. In addition, the conflict between the indigenous population and the mining industry for example regarding human rights can be avoided or reduced in these countries \cite{bridge2004contested}, \cite{kemp2010mining}. Due to the high and heterogeneous mass flows in copper and aluminium production, there is
great interest in the classification of recycling materials in real
time in order to classify the secondary metallic raw materials
according to existing standards and regulations. 

In a digitized recycling process, a precise testing of inho-
mogeneous materials online would allow optimal control
of the composition of the recycled product (alloy-to-alloy
recycling). In particular, more scrap metal could be recycled
to produce high-quality alloys in a targeted manner, making
the production of metal economically and environmentally
more sustainable. According to the state of the art, there is
currently no non-destructive metrological solution for copper
or aluminium production.

Scrap is currently assessed using samples that are viewed as
representative. Processes such as X-ray fluorescence analysis
(XRF), optical emission spectroscopy (OES) or laser-induced
breakdown spectroscopy (LIBS) only allow surface based analysis.
This can only be used for homogeneous materials or requires
complex sample collection and chemical preparation. These
are often neither economically feasible nor can they be carried
out online.

The measuring systems based on PGNAA allow non-destructive elemental analysis of complex material flows or material batches and differ from existing methods in that the material to be measured is analyzed integrally. During a measurement, a neutron-induced gamma spectrum is recorded, the evaluation of which allows the complete elemental composition of the material batch to be determined. With PGNAA, neutrons penetrate the sample material, interact with the nucleii therein and generate atoms in an excited state.
For this work, a demonstrator measurement facility was used, wherein free neutrons are produced using a neutron generator. Monoenergetic neutrons of 2.45 MeV are moderated to thermal energies to maximize the probability of neutron capture within the sample material.
The successive relaxation of the atomic nuclei of the sample material induces a characteristic gamma radiation for the elements (even nuclides) contained in the sample material, which is measured with a detector. 
Subsequently, the elemental composition of large-volume samples and any material flows is analyzed non-
destructively, regardless of their chemical condition, coatings
or impurities. 
This means that inhomogeneous materials can be analyzed without costly sampling and chemical preparation, which are destructive and often can neither be carried out in economical nor online. The measuring method is robust and is used, among other things, in oil exploration, in the detection of explosives and as a online capable method for quality assurance in the cement industry and coal mining \cite{paul2000prompt,glascock2004overview,greenberg2011neutron}. 

Through modern, high-resolution detector systems such as High-purity germanium (HPGe) detectors and newly developed evaluation algorithm softwares
the problem of element identification in the PGNAA spectra can be solved. The software analyzes the peaks in the PGNAA spectrum (Fig. 2) in comparison with nuclear databases \cite{havenith2020quantom}.

The contribution to the improvement of the gamma detectors is done for example by research on the readout electronics of the detectors\cite{WANG_Xiaohu}, \cite{XianqinLi_HaiboYang}.
\\

Classification by spectroscopy is used in various fields such as the differentiation of tablets, juices, wines \cite{jernelv2020convolutional}, the search of iron ore deposits \cite{min8070276}, the deciphering of chemical composition of the atmospheres of extrasolar planets \cite{matchev2022unsupervised}, the detection of radioactive materials \cite{9625926} or for nuclear safety \cite{curtis2019special}. 
Different Spectrometer are used, such as FTIR, NIR, MIR and Raman \cite{jernelv2020convolutional,balytskyi2021raman,liu2017deep}, angle-resolved photoemission spectroscopy (ARPES) \cite{kim2021}, laser-induced breakdown spectroscopy (LIBS) \cite{bhardwaj2021semisupervised} or PGNAA \cite{peng2016research}.

The established analyses in spectral analysis are based on machine learning methods. Neural networks (NNs) were already applied in the 1990s for the identification of isotopes in the PGNAA spectra \cite{olmos1991new,olmos1992application,vigneron1996statistical}. In certain applications such as radioactive material identification, it has been shown that some newly developed machine learning (ML) and deep learning (DL) methods (deep belief network, random forest regression (RFR), logistic regression (LR)) have better performance than the conventional method based on the region of interest (ROI) \cite{9625926,ayhan2021new}. 
Backpropagation neural network (BPNN) is significantly better at recognizing fabrics than Monte Carlo Library Least-Squares (MCLLS) \cite{peng2016research}. It has also been shown that convolutional models should be preferred over fully connected architectures \cite{kamuda2019automated}.
A comparison of convolutional neural network (CNN) to partial least squares-discriminant analysis (PLS-DA), NN, LR, and support vector machine (SVM) was made for NIR-, MIR-, Raman-spectra. CNN generally performed better on raw data than the classification and regression models \cite{jernelv2020convolutional}.
For the classification of nuclear material, the decision tree, random forest, neural network and Bayesian network methods were used. The accuracy of each method was between 98 and 100 percent. However, under certain conditions (a p-value of 1e-09), the decision tree classifier achieved the highest accuracy of 99.6\% \cite{curtis2019special}.

The simulation of the data is done by Monte Carlo methods (MCNPX, MCNP, MCNP4C) \cite{ghal2016quantitative,sahiner2017gamma,doostmohammadi2010combined} a coupled MCNP and GADRASDRF approach \cite{kamuda2019automated}, matrix based method \cite{ryan1995new} and sampling methods for the addition of
noise \cite{kim2021}.
\\

We aim to fill literature gap by presenting the approach
of online classification of metal alloys by PGNAA as a
novelty. This approach has not been used by anyone before.
Furthermore we propose a new access to online classifica-
tion by interpreting a spectrum as a probability distribution.
Specifically, we interpret each measured gamma energy value
as the probability of its occurrence.
Here we do not restrict to
individual peaks or energy ranges, instead we use the complete
energy range and thus all information of the spectrum. For this,
we only need one fully measured spectrum (see Fig. \ref{figure:Spektrum_plot} or \ref{figure:Spektrum_plot_2}) per
material as training data and no data preprocessing. This al-
lows us to use probabilistic methods (kernel density estimator,
maximum log-likelihood method) for classification in order to
significantly and sustainably improve scrap recycling and its
economic and environmental aspects.

For the classification task, a fully measured spectrum is obtained for each material. Using a kernel density estimator, the corresponding probability distribution can be generated for each of the materials' spectra. The kernel density estimator uses the known (stochastic) densities from the data analysis. 
To assign an unknown short time measurement to a material, we use the maximum (log-)likelihood method. The maximum (log-)likelihood method assigns the short time measurement to the most fitting distribution of a fully measured spectrum and thus to the corresponding material.

This method does not reduce the spectra to their individual peaks, but uses the entirety of the measured data. This has the advantage that all details of the data are used for the classification. In this way, even similar copper and aluminium alloys can be classified in under 0.5 second\footnote{All measurement times are stated with respect to the PGNAA demontrator facility used in the research project MetalClass } with a certainty of over 90 $\%$ (cf. Fig. \ref{figure:confusion_cu} and \ref{figure:confusion_al}).

We simulate new spectra by sampling. For that we need only one complete measured spectrum of a material to gain  training, validation and test data. Therefore, any number of spectra can be generated easily and quickly from a single spectrum. Depending on the size of the random samples, any short measurement times can be simulated. This method generates 50 short time spectra per second. This saves the costly acquisition of new data. Furthermore training, validation and test data are very essential for parameter estimation of the kernel density estimator and training of any models such as neural networks.

In contrast to regression, each material class must be clearly
defined by means of a completely measured spectrum of a
material. With the method mentioned, we can carry out a online classification of heterogeneous substances. This means
that similar substances such as aluminium and copper alloys
can in principle be classified online. We test our method
to classify 20 substances. These included five aluminium and
six copper alloys. The presented method scores better than a
convolution neural network (CNN). CNN do not classify short
time spectra (measurement time less than 1 sec) of copper and
aluminium alloys as well compare \cite{cheng2022prompt}. 

The aim of Section \ref{Chapter2:Data} is to introduce the PGNAA and the structure of the data. The methods for classification are presented in Section \ref{chapter3:methods} and  the method for data acquisition introduced in Section \ref{chapter4:simulation}. The optimization of hyperparameter and the results of the classification are dealt with in Section \ref{chapter5:Results}. Finally, Section \ref{Chapter6:conlusion} concludes our findings.
\section{PGNAA \& Data}\label{Chapter2:Data}

Neutron activation analysis (NAA), in which the sample material is bombarded with neutrons, is used to determine the metal compound. There are different types of NAA, such as prompt gamma neutron activation analysis (PGNAA), delayed gamma neutron activation analysis (DGNAA), instrumental neutron activation analysis (INAA), and radiochemical neutron activation analysis (RNAA) \cite{glascock2004overview}. 

\begin{figure}[h]
 \centering
 \includegraphics[scale = 0.198]{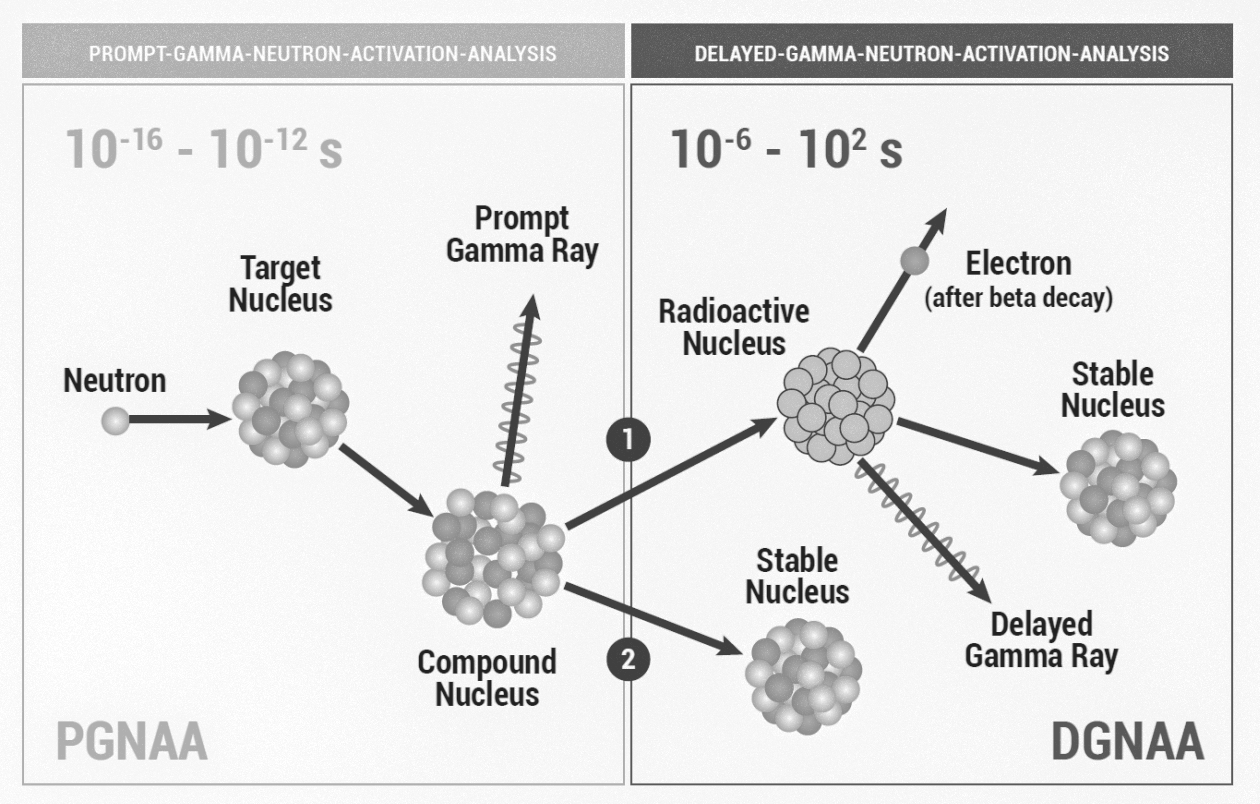}
 \caption{Principle physical mechanisms of P{\&}DGNAA. Source: AiNT GmbH}\label{figure:PGNAA}
\end{figure}

PGNAA analyzes material continuously by irradiating it with a neutron beam. In this process, the atomic nuclei are excited. Various nuclear reactions can be induced by neutron irradiation. The excited nucleus can relax by emitting one or more gamma quanta (see Fig. \ref{figure:PGNAA}, \cite{havenith2020quantom}). 
The excitation of the atomic nuclei of the sample material induces a characteristic gamma radiation for the sample material, which is measured with a gamma ray spectrometer.
The gamma energies are thereby stored in a table (see table \ref{table:Spektrum_tabelle}). 
\begin{table}[!ht]
 \centering
 \caption{Data set of Cu\_1 spectrum.}\label{table:Spektrum_tabelle}
 \includegraphics[scale = 0.2835]{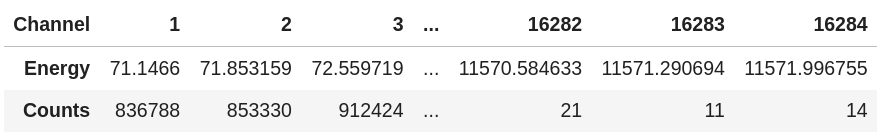}
\end{table} 
The row ''Counts'' gives the amount with which the corresponding energy values were measured.
From Table \ref{table:Spektrum_tabelle} we see the number of channels, the difference between two neighbouring energies (about 0.7 keV), typical numbers of counts and the energy range we measured. We did neither analyse changing external circumstances nor different calibrations, so throughout all datasets the measurement background and energy calibration are kept as fixed as possible.

All PGNAA spectra used in this paper were acquired with an HPGe detector\footnote{You can access online the datasets that were used in this work: \url{https://www.kaggle.com/datasets/smartfactoryowl/metalclass}}. 

\begin{figure}[h]
 \centering
 \includegraphics[scale = 0.227]{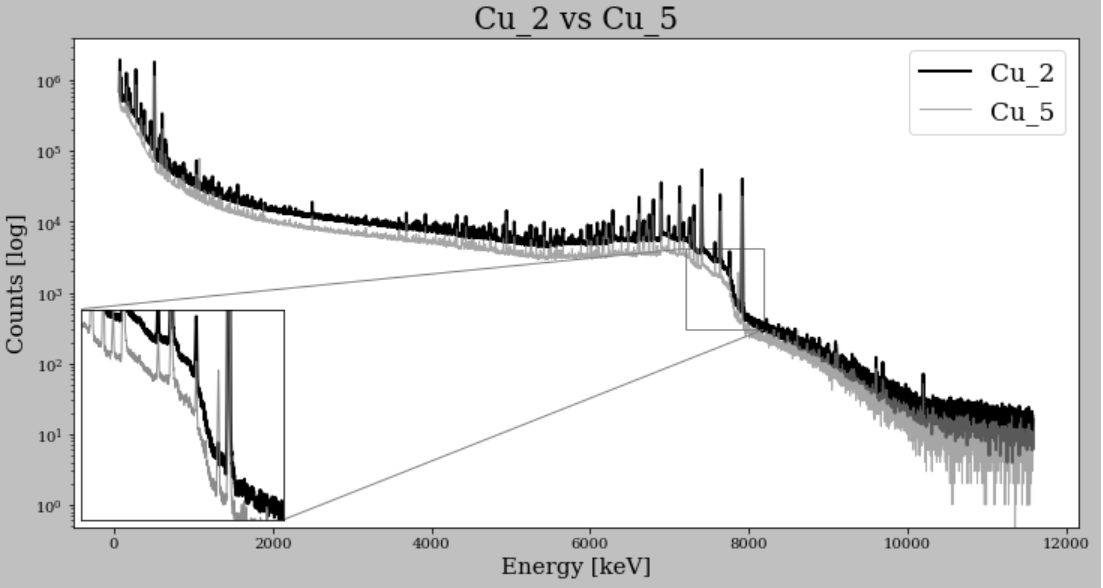}
 \caption{Spectrum of copper alloy Cu\_2 and copper alloy Cu\_5 with each a measurement time of 3 hours.}\label{figure:Spektrum_plot}
\end{figure}

\begin{figure}[h]
 \centering
 \includegraphics[scale = 0.227]{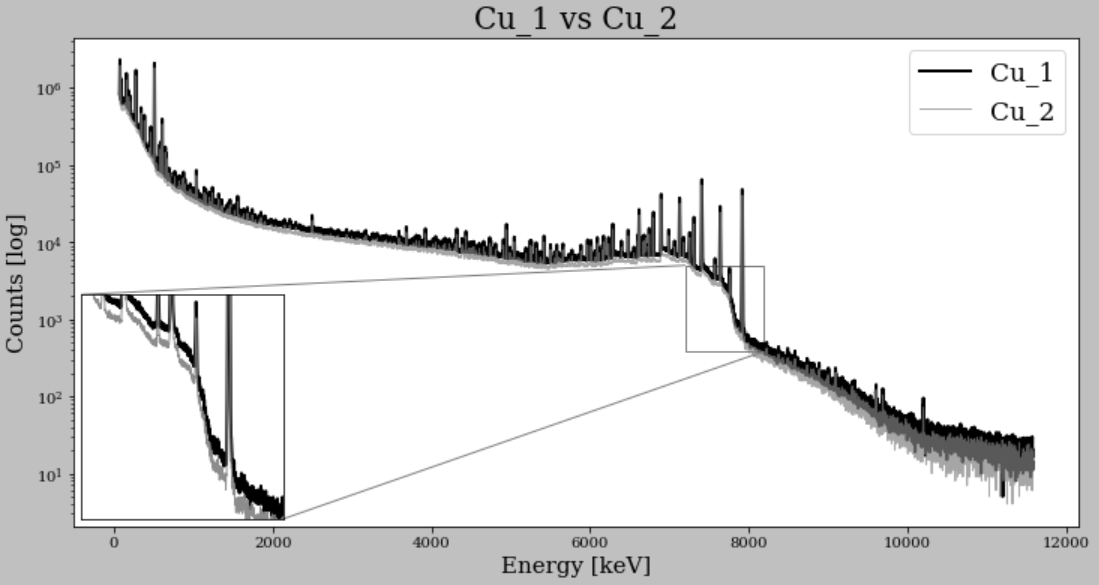}
 \caption{Spectrum of copper alloy Cu\_1 and copper alloy Cu\_2 with each a measurement time of 3 hours. These are spectra of the most similar alloys that will be analyzed.}\label{figure:Spektrum_plot_2}
\end{figure}

The relationship between the energies and the absolute frequencies from the data set (see table \ref{table:Spektrum_tabelle}) can be visualized in a histogram, see for example Fig. \ref{figure:Spektrum_plot} or \ref{figure:Spektrum_plot_2}.
From the figures, the similarity between the alloys and in particular between Cu\_1 and Cu\_2 can be seen.

The zoomed areas in Fig. \ref{figure:Spektrum_plot} and Fig. \ref{figure:Spektrum_plot_2} represent the same areas of the corresponding spectra. The zoomed area in Fig. \ref{figure:Spektrum_plot} shows an additional peak of Cu\_5 compared to Cu\_1, Cu\_2. The alloys from Fig. \ref{figure:Spektrum_plot} and analogy the alloys of Fig. \ref{figure:Spektrum_plot_2} are later used in the classification dependent of time in the same relationship.
Thus, the bigger differences between the spectra of the copper alloys in Fig. \ref{figure:Spektrum_plot} lead to the quicker classification seen between these copper alloys in Fig. \ref{figure:Klassifikation} and the smaller differences between the spectra of the copper alloys in Fig. \ref{figure:Spektrum_plot_2} lead to the much slower classification between these copper alloys seen in Fig. \ref{figure:Klassifikation_1}.
\section{Kernel density estimation \& maximum log-likelihood method}\label{chapter3:methods}
For the classification of the short-time spectra, we use the probability distribution of a single completely measured spectrum. To obtain this distribution, we use the kernel density estimator (see \cite{gramacki2018nonparametric}, chapter 2). The kernel density estimator satisfies the definition of a probability distribution and can be understood as the averaging of density functions concentrated around the individual data points. Based on this distribution, a short-term measurement can be classified using the maximum likelihood method (see \cite{millar2011maximum}, chapter 2.2).

Given a completely measured spectrum with energies $(\gamma_1,\ldots,\gamma_n)$ with $\gamma_i<\gamma_{i+1}$ and corresponding counts of measured photons $(c_1,\ldots, c_n)$ with $c_i\in\mathbb{N}_0$. In our case the materials are always measured fixed at the same energy values. We obtain the corresponding discrete probability distribution $(\hat{f}(\gamma_1),\ldots,\hat{f}(\gamma_n))$ of the energy of a measured photon by using the following kernel density estimator: 
\begin{equation*}
\hat{f}(\gamma_j)=\dfrac{1}{nha}\sum\limits_{i=1}^n c_i k\left(\dfrac{\gamma_j-\gamma_i}{h}\right).
\end{equation*}
Where the kernel $k$ is a non-negative integrable function with integral value 1, $h>0$ is the smoothing factor called bandwidth and $a$ $(a=\sum_{j=1}^n\frac{1}{nh}\sum\limits_{i=1}^n c_i k\left(\frac{\gamma_j-\gamma_i}{h}\right))$ is a normalization factor for the discrete case.
For example, $k(x)=\frac{1}{\sqrt{2\pi}}\exp\left(-\frac{1}{2}x^2\right)$ is a Gaussian kernel and $k(x)=\frac{1}{\pi(1-x^2)}$ a Cauchy kernel (see \cite{gramacki2018nonparametric}, chapter 3). 

The choice of bandwidth and kernel affects the shape of the density obtained. For small bandwidths, the density converges to the discrete estimated distribution.

The maximum log-likelihood method can be used to classify a short term spectrum. It assigns a short term spectrum to the best fitting distribution. 
Let $\mathbb{S}$ be the set of all completely measured spectra and let a probability distribution obtained by kernel density estimation from a spectra $S\in \mathbb{S}$ be given as $(\hat{f}_S(\gamma_1),\ldots,\hat{f}_S(\gamma_n))$. Furthermore, let a short-time measurement $s$ be given with the gamma energies $(\gamma_1,\ldots,\gamma_n)$ and the measurement frequencies $(c_1',\ldots, c_n')$. Evaluation of the short time measurement according to the maximum log-likelihood method for the spectrum $S$ with corresponding probability distribution yields
\begin{equation}\label{equation:likelihood}
\log(p(s|S))=\log\left(\prod_{i=1}^n \hat{f}_S(\gamma_i)^{c_i'}\right)=\sum_{i=1}^n c_i'\log \hat{f}_S(\gamma_i).
\end{equation}
Maximum log-likelihood assigns a short term measurement to the distribution whose likelihood evaluates to the largest value
\begin{equation}\label{equation:max}
\max_{S\in\mathbb{S}} \log(p(s|S)).
\end{equation}
The expression (\ref{equation:max}) can be determined with the help of matrix-vector multiplication, where the matrix and vector entries are $\log\hat{f}_{S_j}(\gamma_i)$ and $c_i'$ respectively.
\section{Data Creation}\label{chapter4:simulation}
Acquiring new data is always associated with high costs and is a necessary component of machine learning (ML). Consequently, we present a methodology for generating training, validation, and test data. We use the obtained data for parameter estimation and neural network training.

\begin{figure}[h]
 \centering
 \includegraphics[scale = 0.1963]{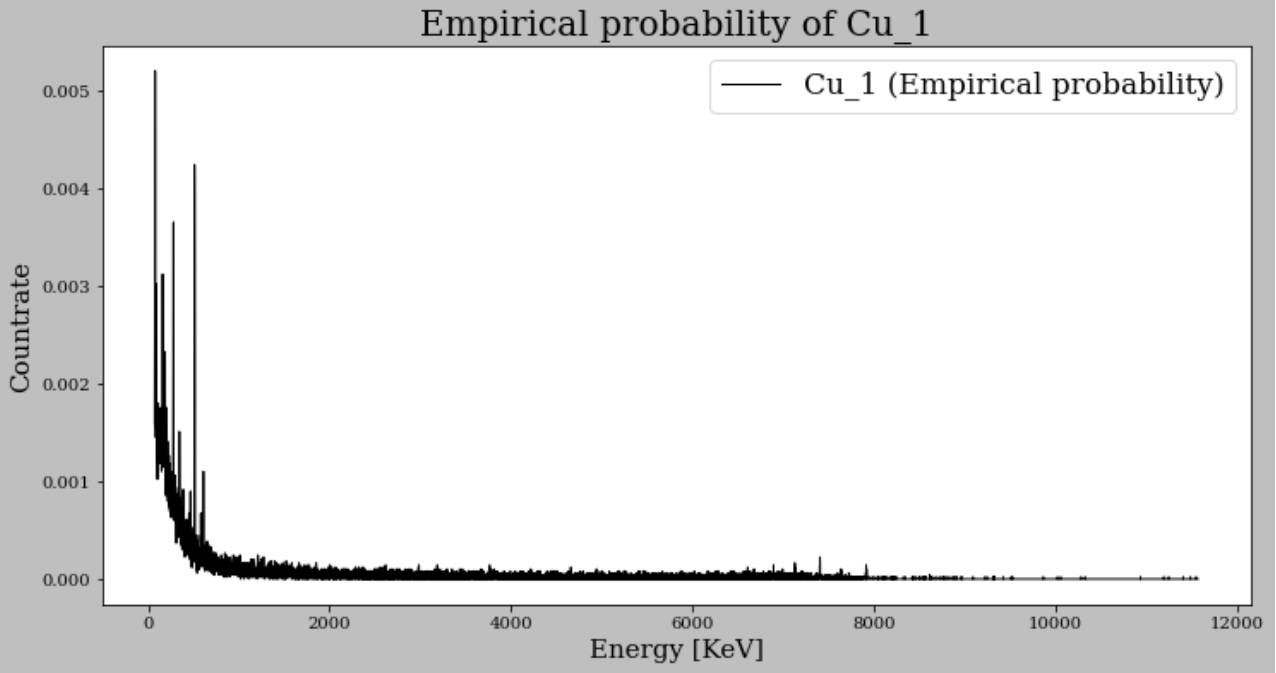}
 \caption{The obtained discrete distribution of the metal alloy Cu\_1.}\label{figure:PGNAA_verteilung}
\end{figure}

\begin{figure}[h]
 \centering
 \includegraphics[scale = 0.23]{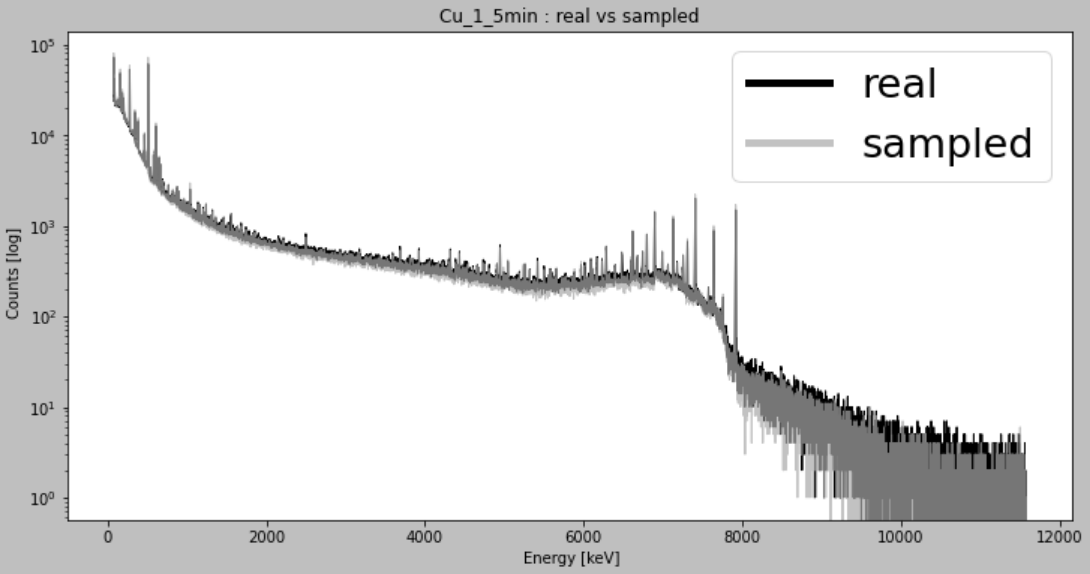}
 \caption{Simulated and real measurement of material Cu\_1 with each a measurement time of 5 minutes.}\label{figure:PGNAA_simulated}
\end{figure}

Since the completely measured spectrum provides us with a representative set of data, we can interpret it as a probability distribution by normalization, which allows the use of probabilistic techniques. With this distribution, we can generate an unlimited amount of new data by sampling as follows:
Let $(\gamma_1,\ldots, \gamma_n)$ be the gamma energies with corresponding
counts of measured photons $(c_1,\ldots, c_n)$. The gamma energies and their count rate should be representative for a completely measured spectrum.
By dividing the count rate $c_i$ by the
total sum $u=(c_1+\ldots+ c_n)$ we obtain the relative
frequencies $(\frac{c_1}{u},\ldots, \frac{c_n}{u})=(p_1,\ldots,p_n)$. Through this procedure, we obtain a discrete distribution of the fully measured spectrum (see Fig. \ref{figure:PGNAA_verteilung}). 
With this distribution, we can now sample the energies by sampling with replacement, by the size and number of times of drawing, we can simulate any measurement time as many times as necessary.
It should be noted that the sum of the absolute frequencies is proportional to the measurement time. 
In our considerations we set 1 sec. as 50000 counts for all materials. We observed this number of counts for our measurement setup in real 1 sec. measurements. Fig. (\ref{figure:PGNAA_simulated}) shows a comparison between a simulated and real spectrum.
\section{Results}\label{chapter5:Results}
\subsection{Choice of the bandwidth}
In the kernel density estimation, the choice of bandwidth and kernel affects the classification quality. To select the optimal parameters we use cross-validation (see more details in \cite{book:2358672}, chapter  11). A parameter is trained on a certain part of the data and tested on the remaining part. This procedure is repeated several times for each parameter and the test results of each parameter are averaged. The averaged results are then compared to select a suitable parameter. For the required training data, we use drawing without replacement.
\begin{figure}[h]
 \centering
 \includegraphics[scale = 0.335]{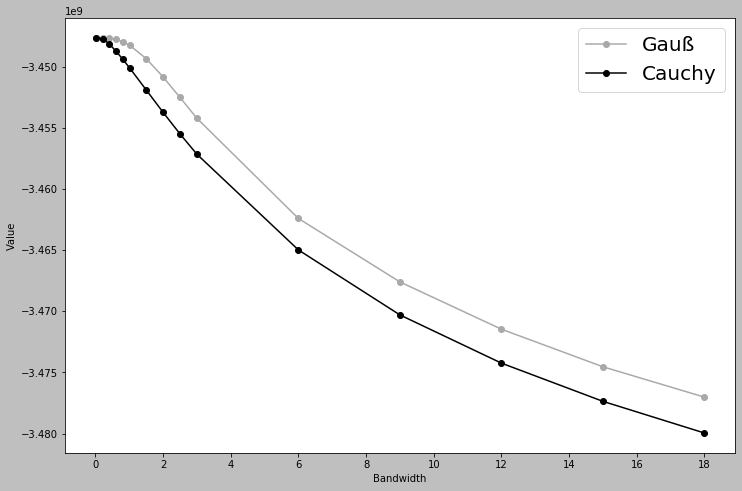}
 \caption{Value of the maximum log-likelihood as a function of the bandwidth. For kernel density, we used Gaussian and Cauchy kernels. Smaller bandwidths maximize the maximium log-likelihood and thus offer themselves as the optimal choice.}\label{figure:Kreuzvalidierung}
\end{figure}
We see that the optimal choice of bandwidth is close to 0, since
higher values lead to better classification and the Cauchy kernel achieves slightly larger values than the Gaussian kernel near 0 (see Fig. \ref{figure:Kreuzvalidierung}). 
Due to the results of Fig. \ref{figure:Kreuzvalidierung}, we perform all our experiments with bandwidth 0.00065 and Cauchy kernel. Compared to the typical line width of the nuclear instrumentation of 2-5 keV FWHM (=full width at half maximum, energy dependent), this bandwidth corresponds to a treatment of the data as discrete values. Using a bandwidth of this range, the Cauchy kernel differs only marginally from the physically expected Gaussian shape. Nevertheless, our tests show slightly better classification performance using a Cauchy kernel for bandwidth near 0.
\subsection{Classification reliability independent of time}
\begin{figure}[h]
 \centering
 \includegraphics[scale = 0.227]{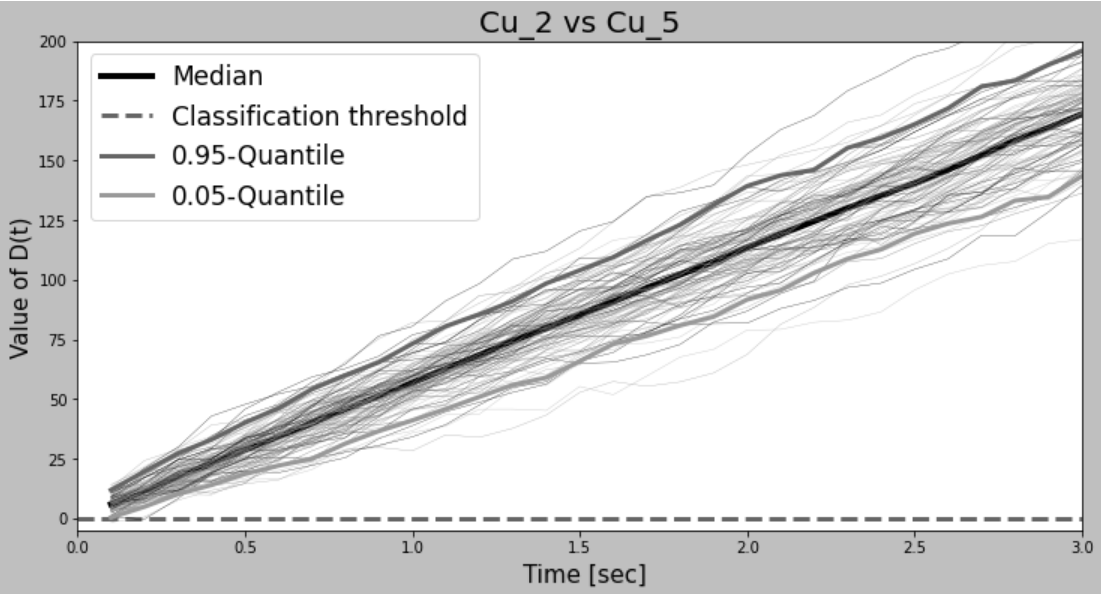}
 \caption{Classification reliability of copper alloys (Cu\_2, Cu\_5) as a function of the measurement time (between 0.03 sec. and 3 sec.) of the spectrum. The longer the measurement time, the more reliable the classification.}\label{figure:Klassifikation}
\end{figure}

\begin{figure}[h]
 \centering
\includegraphics[scale = 0.227]{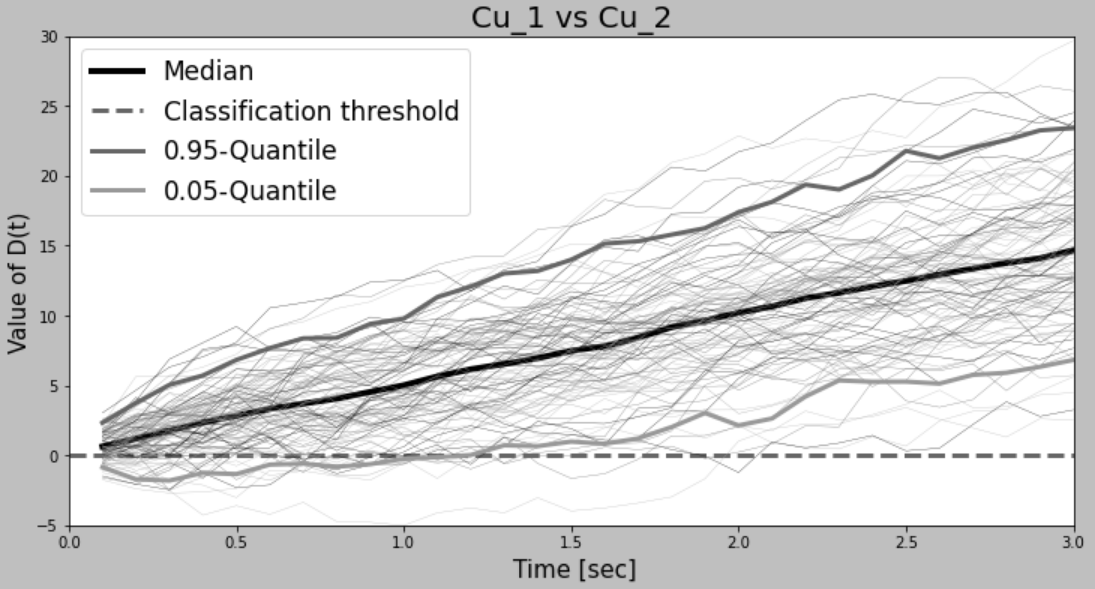}
 \caption{Classification reliability of copper alloys (Cu\_1, Cu\_2) as a function of the measurement time (between 0.03 sec. and 3 sec.) of the spectrum. The most similar alloys represent the worst case in the classification. }\label{figure:Klassifikation_1}
\end{figure}

\begin{figure}[h]
 \includegraphics[scale = 0.226]{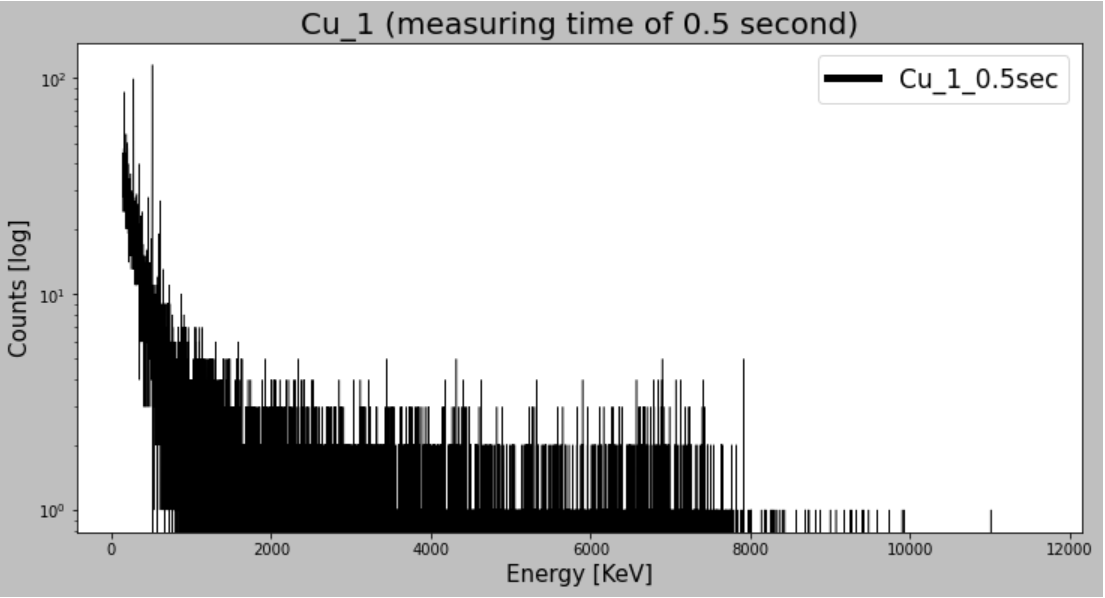}
 \caption{Simuleted spectrum of copper alloy Cu\_1 with a measurement time of 0.5 second.}\label{figure:messzeit_0.5sec}
\end{figure}

As each measured gamma ray yields an additional summand in (\ref{equation:likelihood}), we can visualize the classification between two classes depending on the measurement time of a sample. For this we consider the sample size from the short time measurement $s$ as a function of the measurement time. And we write this as $s(t)$.
For given short time measurement $s(t)$ for two classes $S_k$ and $S_l$ the difference $\log(p(s(t)|S_k))-\log(p(s(t)|S_l))$ can be considered as classification base. In this context we can consider a visualization of the function $D(t)=\log(p(s(t)|S_k))-\log(p(s(t)|S_l))$ and if the curve is below the classification threshold, it is a misclassification.(see Fig. \ref{figure:Klassifikation} and \ref{figure:Klassifikation_1}).
The classification reliability, which increases linearly with the sample size, can be seen in the increasing differences. This shows that the shorter the measurement time of the alloy, the more difficult it is to classify the metal alloys.
This is because the shorter the measurement time, the stronger the noise and the smaller the amount of measurement of individual energies (compare the height of the counts and the noise between Fig. \ref{figure:Spektrum_plot} and \ref{figure:messzeit_0.5sec}).  
The classification of the worst case is shown in Fig. \ref{figure:Klassifikation_1}.

Spectra are required continuously during the irradiation of the sample. Therefore, prompt gamma radiation accounts for the vast majority of the signal. The proportion of delayed gammas in the spectra has not been quantified, but is expected to be more than one order of magnitude lower than prompt gammas. The acquisition times shown here are not comparable to typical measurement times for DGNAA at research reactors, since the neutron flux in the sample is approximately 3-4 orders of magnitude lower for the MetalClass demonstrator facility.
\subsection{Classification evaluation}
A confusion matrix or error matrix (see \cite{rhys2020machine}, chapter 3) is a tool for visualizing and evaluating the predictive performance of a classifier. The accuracy, given as a probability between 0 and 1, can be understood intuitively.

\begin{figure}[h]
 \centering
 \includegraphics[scale = 0.3503]{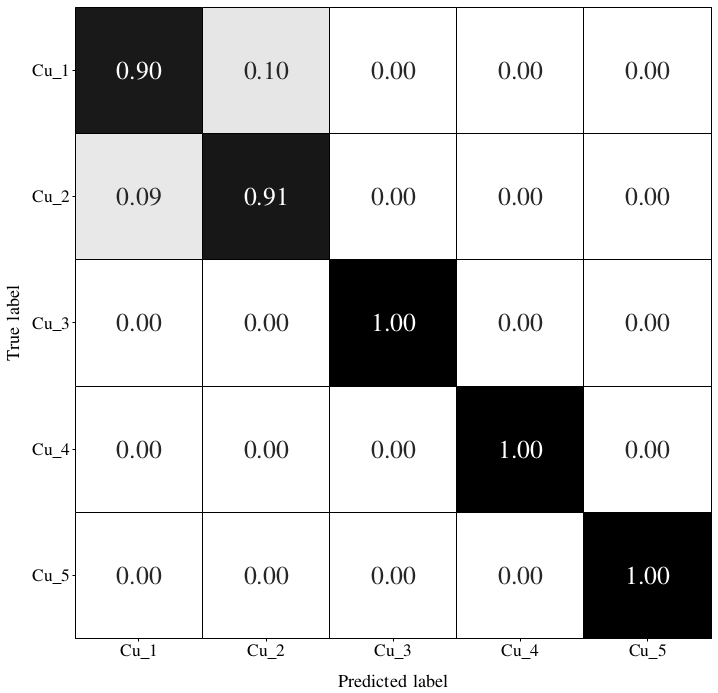}
 \caption{Confusion matrix for different copper alloys.}\label{figure:confusion_cu}
\end{figure}

The following confusion matrix (see Fig. \ref{figure:confusion_cu})
shows classification results of five copper alloys (Cu\_1, Cu\_2, Cu\_3, Cu\_4, Cu\_5). 1000 test data per alloy with a measuring time of 0.5 seconds were prepared for the classification. The bandwidth 0.00065 and Cauchy density were chosen for the Kernel density estimation. The results clearly show that alloy Cu\_1 and Cu\_2 can be distinguished from each other slightly worse than the rest of the copper alloys. 

\begin{figure}[h]
 \centering
 \includegraphics[scale = 0.3503]{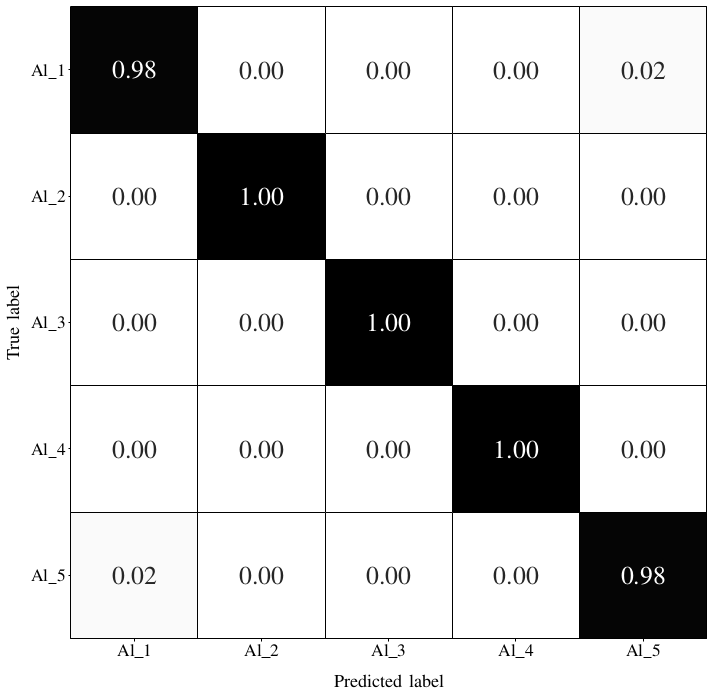}
 \caption{Confusion matrix for different aluminium alloys.}\label{figure:confusion_al}
\end{figure}

The next confusion matrix (see Fig. \ref{figure:confusion_al}) 
presents the classification results of aluminium alloys (Al\_1, Al\_2, Al\_3, Al\_4, Al\_5). For aluminium, the measurement time was set at 0.25 seconds. In this case, all five alloys have approximately the same degree of difficulty in distinction.

It should be noted that all the alloys studied, with the exception
of the copper alloys Cu\_1 and Cu\_2, can be classified almost
100 \% in a time of less than 0.25 seconds. 

\begin{figure}[h]
 \includegraphics[scale = 0.287]{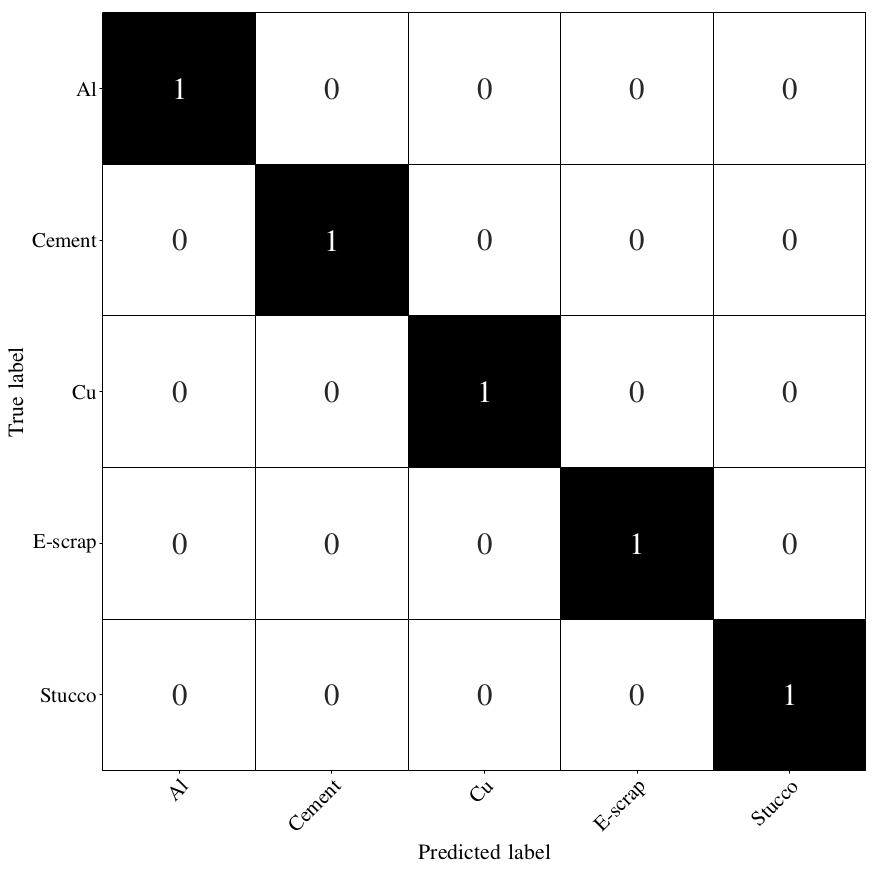}
 \caption{Confusion matrix for different materials.}\label{figure:confusion_alle}
\end{figure}

To illustrate that different materials can be classified very easily, we reduce the measurement time of the test data to 0.0625 seconds and the Fig. \ref{figure:confusion_alle} is the comparison of different materials (such as aluminium, cement, copper, E-scrap and stucco). We also classified soil, batteries, ore among themselves and melamine, PVC, ASILIKOS among themselves and were also able to classify 100 \% correctly at the same time (0.0625 sec.).

Due to the binning of the existing detector channels, a slight drop in the classification rate can be observed with different measuring times of spectra. For example, after quartering the number of channels by adding the energies of the corresponding channels for the five copper alloys with a measurement time of 0.25 seconds and a bandwidth of 0.00065, the accuracy decreases by about 1.8 \%. 
\section{Conclusion}\label{Chapter6:conlusion}
The goal of optimizing recycling processes for aluminium and copper alloys through online classification can be achieved using the kernel density estimator and the maximum likelihood method. For example, most alloys can be classified close to 100 \% at 0.25 seconds. We can classify faster than a CNN and only need a single fully measured spectrum of the material to be classified by using the full information of the spectrum. We can classify online without data preprocessing and without additional training data. Nevertheless we have introduced a simple and fast method for simulating an unlimited number of different short- and long-term spectra. In doing so, we can generate 50 spectra under 1 sec. This can be of great advantage for various methods and areas of AI such as comparing, training, validating and optimizing models.

It remains to be investigated whether the classification can be further optimized by feature engineering, since the measurement method and spectrum have many physical features. Alternative detector types with lower resolution but faster signal acquisition may also become interesting due to the fast detection. 
\section{Acknowledgments}
The MetalClass project is funded by the German Federal Ministry of Education and Research with the funding reference number 01IS20082A/B.
The copper alloys were provided by Wieland-Werke AG. 

The authors would like to thank Dr. Stefan Theobald and Sebastian Bender for support and discussions.
\bibliographystyle{IEEEtran}
\bibliography{bibmetal.bib}

\end{document}